\pdfoutput=1

\documentclass[11pt]{article}

\usepackage[final]{acl}

\usepackage{times}
\usepackage{latexsym}

\usepackage[T1]{fontenc}

\usepackage[utf8]{inputenc}

\usepackage{microtype}

\usepackage{inconsolata}

\usepackage{graphicx}

\usepackage[edges]{forest}
\usepackage{booktabs}
\usepackage{enumitem}
\usepackage{bbding}
\usepackage{array}
\usepackage{float}
\usepackage{placeins}

\newcommand{\eat}[1]{}
\newcommand{\spara}[1]{\smallskip\noindent{\bf #1}}

\title{Large Language Models Meet Knowledge Graphs for Question Answering: Synthesis and Opportunities}
\author{
 \textbf{Chuangtao Ma\textsuperscript{1}},
 \textbf{Yongrui Chen\textsuperscript{2}},
 \textbf{Tianxing Wu\textsuperscript{2}}\thanks{Corresponding authors.},
 \textbf{Arijit Khan\textsuperscript{3, 1}}\footnotemark[1],
 \textbf{Haofen Wang\textsuperscript{4}}
\\
 \textsuperscript{1}Aalborg University, Denmark,
 \textsuperscript{2}Southeast University, China \\
 \textsuperscript{3}Bowling Green State University, USA,
 \textsuperscript{4}Tongji University, China
\\
\small{
\texttt{chuma@cs.aau.dk, \{yongruichen, tianxingwu\}@seu.edu.cn, arijitk@bgsu.edu, haofenwang@tongji.edu.cn}}
}

\begin{document}

\maketitle

\begin{abstract}
Large language models (LLMs) have demonstrated remarkable performance on question-answering (QA) tasks because of their superior capabilities in natural language understanding and generation. However, LLM-based QA struggles with complex QA tasks due to poor reasoning capacity, outdated knowledge, and hallucinations.
Several recent works synthesize LLMs and knowledge graphs (KGs) for QA to address the above challenges. 
In this survey, we propose a new structured taxonomy that categorizes the methodology of synthesizing LLMs and KGs for QA according to the categories of QA and the KG's role when integrating with LLMs. 
We systematically survey state-of-the-art methods in synthesizing LLMs and KGs for QA and compare and analyze these approaches in terms of strength, limitations, and KG requirements. 
We then align the approaches with QA and discuss how these approaches address the main challenges of different complex QA. 
Finally, we summarize the advancements, evaluation metrics, and benchmark datasets and highlight open challenges and opportunities. 
\end{abstract}

\section{Introduction}

Question answering (QA) plays a fundamental role in artificial intelligence, natural language processing, information retrieval, and data management areas since it has a wide range of applications, such as text generation, chatbots, dialog generation, web search, entity linking, natural language query, fact-checking, etc. 
The pre-trained language models (PLMs) and recent LLMs have shown superior performance in several QA tasks such as KBQA (Knowledge bases QA), KGQA (Knowledge graph QA), CDQA (Closed domain QA), etc. 
However, PLM and LLM-based methods are incapable of handling complex QA scenarios due to the following limitations. {\bf(1)} {\em Limited complex reasoning capability:} LLMs encapsulate very limited reasoning capability since they have been pre-trained with the tasks of predicting the next word in a text sequence.  {\bf(2)} {\em Lack of up-to-date and domain-specific knowledge:} LLMs are incapable of generating accurate and up-to-date responses for domain-specific QA because they have been pre-trained on the data and knowledge with a cutoff date. {\bf(3)} {\em Tendency to generate hallucinated content:} LLMs usually generate hallucinated content due to the lack of factual verification and logical consistency checking. 

\vspace{-1mm}
\spara{Challenges.} Retrieval augmented generation (RAG)~\citep{mao2021generation} was proposed for open-domain QA by retrieving the relevant contexts from large documents, and several techniques, such as graph neural networks (GNNs)~\citep{li2025graph}, have been investigated to enhance the retrieval coverage from passages.
Although RAG-based QA can generate better responses in comparison to NoRAG-based QA, it still has limited capability for knowledge reasoning and understanding user interactions during complex QA. 
Complex QA usually involves knowledge interactions and fusion among data across modalities and sources, and an excellent understanding of complex queries and user interactions, whereas the RAG-based QA suffers a lot from the following technical challenges.  {\bf(1)} {\em Knowledge conflicts:} Conflicts occur due to the fusion of inconsistent and overlapping knowledge between LLMs and external sources in RAG-based QA that may further tend to generate inconsistent answers. {\bf(2)} {\em Poor relevance and quality of retrieved context:} The accuracy of the generated answers in RAG-based QA largely depends on the relevance and quality of the retrieved context, where irrelevant context leads to incorrect results.  {\bf(3)} {\em Lack of iterative and multi-hop reasoning:} RAG-based QA struggles to generate accurate and explainable answers for questions requiring global and summarized contexts due to a lack of iterative.

The emergence of synthesizing LLMs+KGs provides a unique opportunity to address the above challenges and limitations of LLMs for knowledge-intensive tasks, e.g., complex QA~\citep{ma2025unifying}.
The graph retrieval augmented generation (GraphRAG)~\citep{zhang2025survey, peng2024graph, han2024retrieval} and knowledge graph retrieval augmented generation (KG-RAG) ~\citep{sanmartin2024kg, yang2024kg} have demonstrated the strengths of unifying LLMs with KGs for complex QA. 
These GraphRAG and KG-RAG based QA approaches introduce several modules, such as knowledge integration and fusion, reasoning guidelines, and knowledge validation and refinement, to mitigate the above challenges.

\spara{Motivation.} In recent years, a rapidly increasing number of works on synthesizing LLMs and KGs for QA have been conducted to achieve complex QA under open-domain and long-context settings. 
This survey aims to address the aforementioned limitations by outlining the recent progress in integrating LLMs with KGs for complex QA, summarizing technical advances, and identifying open challenges and future research opportunities.
Our survey differs from the existing surveys (\textbf{details in Appendix \S \ref{subsec:appendix-related-survey}}), it provides a comprehensive overview of the recent advancements of QA from the perspective of the roles of KGs when synthesizing LLMs and KGs for complex QA. 

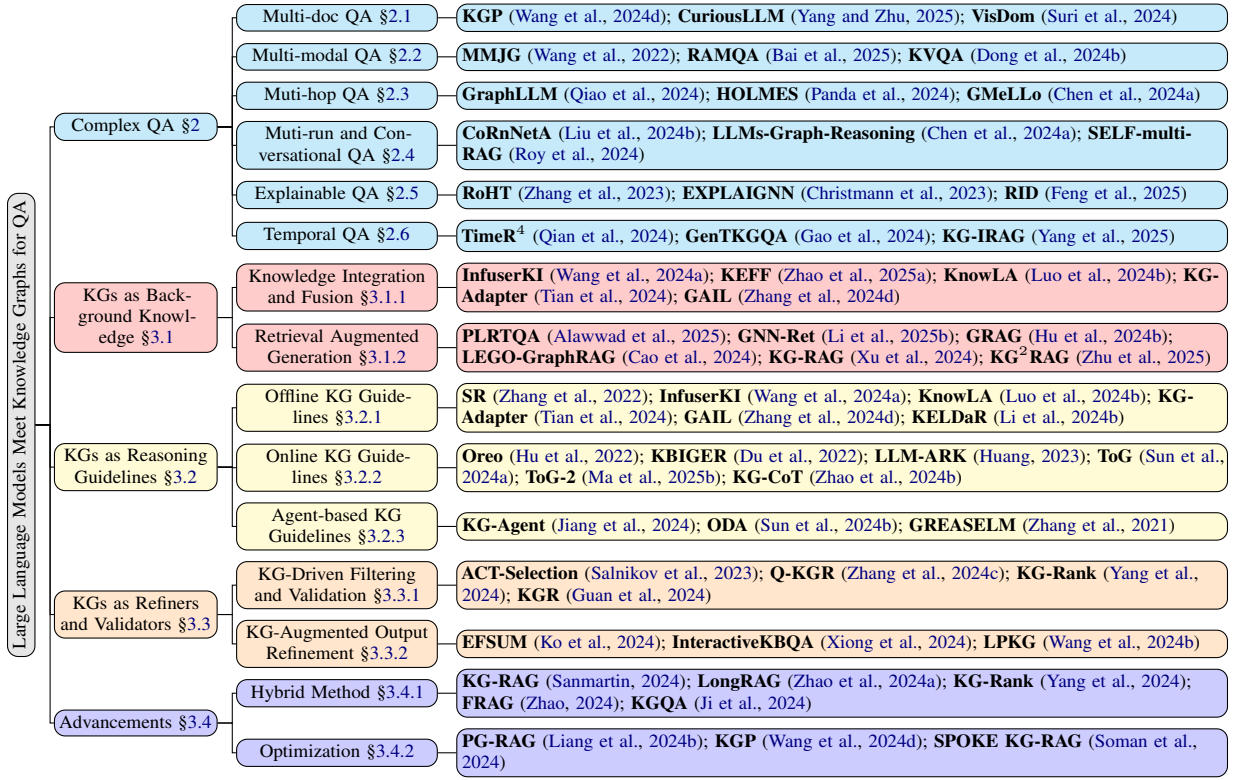
\begin{figure*}[ht]
	\centering
        \tiny
	\begin{forest}
		for tree={
			forked edges,
			grow'=0,
			draw,
			rounded corners,
			node options={align=center},
			calign=edge midpoint,
            font=\scriptsize,
		},
		[Large Language Models Meet Knowledge Graphs for QA, rotate=90, text width=6cm, fill=black!10
			[Complex QA §\ref{sec:complexqa}, text width=2.0cm, for tree={fill=cyan!20}
                    [Multi-doc QA  §\ref{subsec:multi-doc}, text width=2.5cm
                        [
                        \textbf{KGP}~\citep{wang2024knowledge};
                        \textbf{CuriousLLM}~\citep{yang2025curiousllm};
                        \textbf{VisDom}~\citep{suri2024visdom},
                        text width=10.0cm, node options={align=left}
                        ]
                    ]
                    [Multi-modal QA  §\ref{subsec:multi-modal}, text width=2.5cm
                        [
                        \textbf{MMJG}~\citep{wang2022knowledge};
                        \textbf{RAMQA}~\citep{bai2025ramqa};
                        \textbf{KVQA}~\citep{dong2024modality},
                        text width=10.0cm, node options={align=left}
                        ]
                    ]
                    [Muti-hop QA  §\ref{subsec:multi-hop}, text width=2.5cm
                        [
                        \textbf{GraphLLM}~\citep{qiao2024graphllm};
                        \textbf{HOLMES}~\citep{panda2024holmes};
                        \textbf{GMeLLo}~\citep{chen2024llm},
                        text width=10.0cm, node options={align=left}
                        ]
                    ]
                    [Conversational QA  §\ref{subsec:conversational}, text width=2.5cm
                        [
                        \textbf{CoRnNetA}~\citep{liu2024conversational};
                        \textbf{LLMs-Graph-Reasoning}~\citep{chen2024llm};
                        \textbf{SELF-multi-RAG}~\citep{roy2024learning},
                        text width=10.0cm, node options={align=left}
                        ]
                    ]
                    [Explainable QA §\ref{subsec:x-qa}, text width=2.5cm
                        [
                        \textbf{RoHT}~\citep{zhang2023reasoning};
                        \textbf{EXPLAIGNN}~\citep{christmann2023explainable};
                        \textbf{RID}~\citep{feng2025retrieval},
                        text width=10.0cm, node options={align=left}
                        ]
                    ]
                    [Temporal QA §\ref{subsec:tem-qa}, text width=2.5cm
                        [
                        \textbf{TimeR$^4$}~\citep{qian2024timer4};
                        \textbf{GenTKGQA}~\citep{gao2024two};
                        \textbf{KG-IRAG}~\citep{yang2025kg},
                        text width=10.0cm, node options={align=left}
                        ]
                        ]
			]
            [KGs as Background Knowledge §\ref{sec:kgasbk}, text width=2.0cm, for tree={fill=red!20}
                [Knowledge Integration and Fusion §\ref{subsec:kgasb1}, text width=2.5cm
                    [
                    \textbf{InfuserKI}~\citep{wang2024infuserki};
                    \textbf{KEFF}~\citep{zhao2025improving}; 
                    \textbf{KnowLA}~\citep{luo2024knowla};
                    \textbf{KG-Adapter}~\citep{tian2024kg};
                    \textbf{GAIL}~\citep{zhang2024gail}, 
                    text width=10.0cm, node options={align=left}
                    ]
                ]
                [Retrieval Augmented Generation §\ref{subsec:kgasb3}, text width=2.5cm
                    [
                    \textbf{PLRTQA}~\citep{alawwad2024enhancing};
                    \textbf{GNN-Ret}~\citep{li2025graph};
                    \textbf{GRAG}~\citep{hu2024grag};
                    \textbf{LEGO-GraphRAG}~\citep{cao2024lego};
                    \textbf{KG-RAG}~\citep{xu2024retrieval};
                    \textbf{KG$^2$RAG}~\citep{zhu2025knowledge},
                    text width=10.0cm, node options={align=left}
                    ]
                ]			
			]
			[KGs as Reasoning Guidelines §\ref{sec:kgasrg}, text width=2.0cm, for tree={fill=yellow!20}
                [Offline KG Guidelines §\ref{subsec:kgasrg1}, text width=2.5cm
                    [
                    \textbf{SR}~\citep{zhang2022subgraph};
                    \textbf{InfuserKI}~\citep{wang2024infuserki};
                    \textbf{KnowLA}~\citep{luo2024knowla};
                    \textbf{KG-Adapter}~\citep{tian2024kg};
                    \textbf{GAIL}~\citep{zhang2024gail};
                    \textbf{KELDaR}~\citep{li2024framework},
                    text width=10.0cm, node options={align=left}
                    ]
                ]
                [Online KG Guidelines §\ref{subsec:kgasrg2}, text width=2.5cm
                    [
                     \textbf{Oreo}~\citep{hu2022empowering};
                     \textbf{KBIGER}~\citep{du2022knowledge};
                     \textbf{LLM-ARK}~\citep{huang2023llm};
                     \textbf{ToG}~\citep{sun2023think, ma2025think};
                     \textbf{KG-CoT}~\citep{zhao2024kg};
                     \textbf{HippoRAG}~\citep{jimenez2024hipporag},
                    text width=10.0cm, node options={align=left}
                    ]
                ]
                [Agent-based KG Guidelines §\ref{subsec:kgasrg3}, text width=2.5cm
                    [
                     \textbf{KG-Agent}~\citep{jiang2024kg};
                     \textbf{ODA}~\citep{sun2024oda};
                     \textbf{GREASELM}~\citep{zhang2022greaselm};
                     \textbf{PoG}~\citep{chen2024plan};
                     \textbf{ATOMR}~\citep{xin2025atomr},
                    text width=10.0cm, node options={align=left}
                    ]
                ]
			]
            [KGs as Refiners and Validators §\ref{sec:kgasrv}, text width=2.0cm, for tree={fill=orange!20}
                [KG-Driven Filtering and Validation §\ref{subsec:kgasrv1}, text width=2.5cm
                    [
                    \textbf{ACT-Selection}~\citep{salnikov2023answer};
                    \textbf{Q-KGR}~\citep{zhang2024question};
                    \textbf{KG-Rank}~\citep{yang2024kg};
                    \textbf{KGR}~\citep{guan2024mitigating},
                    text width=10.0cm, node options={align=left}
                    ]
                ]
			[KG-Augmented Output Refinement §\ref{subsec:kgasrv2}, text width=2.5cm
                    [
                    \textbf{EFSUM}~\citep{ko2024evidence};
                    \textbf{InteractiveKBQA}~\citep{xiong2024interactive};
                    \textbf{LPKG}~\citep{wang2024learning},
                    text width=10.0cm, node options={align=left}
                    ]
                ]
			]
			[Advancements §\ref{sec:advance}, text width=2.0cm, for tree={fill=blue!20}
            	[Hybrid Method §\ref{subsec:hybridm}, text width=2.5cm, for tree={fill=blue!20}
                       [
                        \textbf{KG-RAG}~\citep{sanmartin2024kg};
                        \textbf{LongRAG}~\citep{zhao2024longrag};
                        \textbf{KG-Rank}~\citep{yang2024kg};
                        \textbf{FRAG}~\citep{zhao2024frag};
                        \textbf{KGQA}~\citep{ji2024retrieval}, 
                        text width=10.0cm, node options={align=left}
                        ]
			    ]
                [Optimization §\ref{subsec:optimization}, text width=2.5cm, for tree={fill=blue!20}
                       [
                        \textbf{PG-RAG}~\citep{liang2024empowering};
                        \textbf{KGP}~\citep{wang2024knowledge};
                        \textbf{SPOKE KG-RAG}~\citep{soman2024biomedical},
                        text width=10.0cm, node options={align=left}
                        ]
			    ]
		   ]
		]
	\end{forest}
    \vspace{0.1mm}
\caption{A Structured Taxonomy of Synthesizing LLMs and KGs for QA. }
    \label{fig:taxonomy}
    \vspace{-3mm}
\end{figure*}

\spara{Taxonmy.} We categorize the methodology of synthesizing LLMs and KGs and complex QA from different perspectives, and a structured taxonomy (\textbf{details in Appendix \S ~\ref{sec:appendix-taxonomy}}) is given in Figure~\ref{fig:taxonomy}. 
The taxonomy from different perspectives aims to highlight the alignments between various LLM+KG approaches and different complex QA by discussing how the LLM+KG approaches with different roles of KG can address the challenges of complex QA.
Notably, these categories, divided from different perspectives, are non-exclusive, where a work might be classified into multiple categories from different perspectives.

\spara{Contributions.} Considering the popularity and mainstream adoption of both LLMs and KGs, and their wide applications in QA, our survey is timely. 
The contributions of this work are summarized below.
\textbf{(1)} We introduce the structured taxonomy that categorizes state-of-the-art (SOTA) works on synthesizing LLMs+KGs for QA. \textbf{(2)} We review the SOTA works on synthesizing LLMs and KGs for QA in various categories and discuss the recent advanced topics in this field. \textbf{(3)} We create the alignments between various approaches in synthesizing LLMs and KGs and complex QA, and highlight how these approaches address the specific challenges of different complex QA; \textbf{(4)} We discuss the current challenges and opportunities in synthesizing LLMs and KGs for QA. 
The online resources of this survey are available on Github\footnote{\url{https://github.com/machuangtao/LLM-KG4QA}}.

\section{Complex QA} \label{sec:complexqa}
Complex QA usually involves question decomposition, knowledge fusion among data across modalities and sources, where complex knowledge reasoning is required to generate accurate answers.\eat{~\citep{lin2025explore}.} The methodology in synthesizing LLMs and KGs for complex QA has been exploited as follows. 

\subsection{Multi-document QA} \label{subsec:multi-doc}
Multi-document QA refers to the QA over contexts from multiple documents, while efficiently and effectively retrieving the relevant knowledge from multiple contexts is the main technical challenge. 
To reduce the retrieval latency and improve the quality of the retrieved context for multi-document QA, KGP~\citep{wang2024knowledge} introduces an LLM-based graph traversal agent for retrieving relevant knowledge from KG. 
Similarly, CuriousLLM~\citep{yang2025curiousllm} integrates a knowledge graph prompting, reasoning-infused LLM agent, and graph traversal agent to augment LLMs for multi-document QA. 
VisDom~\citep{suri2024visdom} introduces a novel multimodal RAG for multi-document question answering by integrating and fusing the multi-modal knowledge and leveraging the (Chain-of-thought) CoT-based reasoning. 

\subsection{Multi-modal QA} \label{subsec:multi-modal}
Multi-modal QA refers to the QA over multi-modal data, and visual QA (VQA) is one of the typical multi-modal QA. 
To retrieve the most relevant knowledge from the external KG for enhancing VQA, MMJG~\citep{wang2022knowledge} introduces an adaptive knowledge selection to jointly select knowledge from visual and text knowledge based on the knowledge-aware attention and multi-modal guidance.
To effectively retrieve the evidence from multi-modal data, RAMQA~\citep{bai2025ramqa} enhances multi-modal retrieval-augmented QA by integrating learning-to-rank with training of generative models via multi-task learning.
KVQA~\citep{dong2024modality} integrates LLMs with multimodal knowledge by using a two-stage prompting and a pseudo-siamese graph medium fusion to balance intra-modal and inter-modal reasoning. 

\subsection{Multi-hop QA} \label{subsec:multi-hop}
Multi-hop QA differs from simple QA, which usually involves multi-step reasoning to generate the final answers. 
The basic idea is to decompose the multi-hop questions into multiple single-hop questions, then generate the answers for each single-hop question, and finally integrate them~\citep{linders2025knowledge}.
For instance, GraphLLM~\citep{qiao2024graphllm} leverages LLMs to decompose the multi-hop question into several simple sub-questions and retrieve the sub-graphs via GNNs and LLMs to generate the answers for sub-questions based on graph reasoning.
HOLME~\citep{panda2024holmes} utilizes a context-aware retrieved and pruned hyper-relational KG that is constructed based on the entity-document graph to enhance LLMs for generating the answers of multi-hop QA.
To enable accurate factual knowledge retrieval and reasoning of LLMs for multi-hop QA, GMeLLo~\citep{chen2024llm} effectively integrates the explicit knowledge of KGs with the linguistic knowledge of LLMs by introducing the fact triple extraction, relation chain extraction, and query and answer generation.  

\subsection{Conversational QA} \label{subsec:conversational}
The challenge of conversational QA lies in how to make the language model (LM) easily understand the questions and intermediate interactions. 
To make user interactions easily understood by machines, CoRnNetA~\citep{liu2024conversational} introduces an LLM-based question reformulation, reinforcement learning agent, and soft reward mechanism to improve the interpretation of multi-turn interactions with KGs.
The conversational QA involves several multi-run QA to refine and get accurate answers through multiple rounds of interactions.
The knowledge aggregation module and graph reasoning are introduced for joint reasoning between the graph and LLMs~\citep{jain2024integrating} to address the challenges of understanding the question and context for conversational QA.
To improve the contextual understanding and the answer quality for conversational QA, SELF-multi-RAG~\citep{roy2024learning} leverages LLMs to retrieve from the summarized conversational history and reuse the retrieved knowledge for augmentation.

\subsection{Explainable QA} \label{subsec:x-qa}
Explainable QA (XQA) aims to provide explanations for the generated answers based on the reasoning over the factual KGs. 
To effectively integrate the multiple sources of knowledge for XQA, RoHT~\citep{zhang2023reasoning} introduces a two-stage XQA method that implements the probabilistic reasoning based on the constructed Hierarchical Question Decomposition Tree (HQDT) from the aggregated knowledge.
To trace the provenance and improve the explainability of the answers, EXPLAIGNN~\citep{christmann2023explainable} constructs a heterogeneous graph from retrieved knowledge and user explanations and generates explanatory evidence based on GNN with question-level attention.
RID~\citep{feng2025retrieval} directly integrates the unsupervised retrieval with LLMs\eat{the decoding of LLMs by dynamically adjusting decoding granularity and direction for adaptively generating the explainable answers} based on reinforcement learning-driven knowledge distillation.  

\subsection{Temporal QA} \label{subsec:tem-qa}
The challenges of temporal QA lie in fully understanding the implicit time constraints and effectively incorporating them with temporal knowledge for temporal reasoning. 
To improve the accuracy of LLMs in answering temporal questions, TimeR$^4$~\citep{qian2024timer4} introduces a Retrieve-Retriev-Rerank pipeline to augment the temporal reasoning of LLMs\eat{with temporal knowledge} by temporal knowledge-based fine-tuning. 
Similarly, GenTKGQA~\citep{gao2024two} introduces a temporal GNN and virtual knowledge indicators to capture\eat{the correlation of neighboring nodes with} temporal knowledge embeddings, and further dynamically integrates retrieved subgraphs into LLMs for temporal reasoning.
To facilitate reasoning of LLMs with\eat{knowledge retrieval from} KGs, KG-IRAG~\citep{yang2025kg} enables LLMs to incrementally retrieve knowledge and evaluate its sufficiency for augmenting the capabilities of LLMs in answering time-sensitive and event-based queries involving temporal dependencies.

\section{Approaches and Alignments} \label{sec:approaches-alignment}
The strengths and limitations of the approach in synthesizing LLMs and KGs for QA are summarized (\textbf{details in Table~\ref{tab:taxonomy-comparsion}}), and the research progress on the alignment of these approaches with the complex QA is discussed (\textbf{details in Table~\ref{tab:alignment-approaches}}).

\subsection{KGs as Background Knowledge} \label{sec:kgasbk}
KGs usually play the role of background knowledge when synthesizing LLMs for complex QA, where knowledge fusion and RAG are the main technical paradigms (\textbf{comparison in Table~\ref{tab:summary-comparsion-kg-as-bgk}}).
\subsubsection{Knowledge Integration and Fusion} \label{subsec:kgasb1}
Knowledge integration and fusion aim to enhance language models (LMs) by integrating unknown knowledge into LMs for \eat{knowledge-intensive tasks, such as} QA. 
\begin{figure}[ht]
    \centering
    \includegraphics[width=1.0 \linewidth]{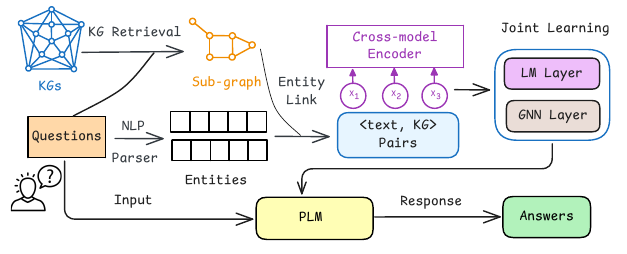}
    \caption{Knowledge Integration and Fusion.}
    \label{fig:1}
\end{figure}
As shown in Figure~\ref{fig:1}, the KGs and text are aligned via local subgraph extraction and entity linking, and then fed into the cross-model encoder to bidirectionally fuse text and KG to jointly train the language models for complex QA tasks~\citep{yasunaga2022deep,zhang2022greaselm}.   
To address knowledge forgetting and noisy knowledge during knowledge integration, InfuserKI~\citep{wang2024infuserki} and KEFF~\citep{zhao2025improving} introduce the adaptive selection and knowledge enhancement filter, respectively, which selects the new knowledge and integrates it with LLMs. 
Fine-tuning LLMs with text and knowledge graphs can improve their performance on specified tasks. 
For instance, KG-Adapter~\citep{tian2024kg} improves parameter-efficient fine-tuning of LLMs by introducing a knowledge adaptation layer to LLMs.
GAIL~\citep{zhang2024gail} fine-tunes LLMs for lightweight KGQA models based on retrieved SPARQL-question pairs from KGs.

\subsubsection{Retrieval Augmented Generation} \label{subsec:kgasb3}
\begin{figure}[ht]
    \centering
    \includegraphics[width=1.0 \linewidth]{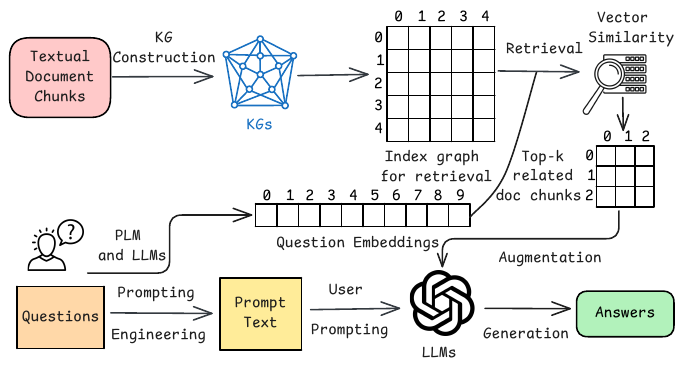}
    \caption{Retrieval Augmented Generation.}
    \label{fig:3}
\end{figure}

RAG serves as a retrieval and augmentation mechanism, as shown in Figure~\ref{fig:3}. It first retrieves relevant knowledge from the text chunks based on vector-similarity retrieval, and then augments the LLMs by integrating the retrieved context with LLMs~\citep{alawwad2024enhancing, li2025graph}.
However, the mainstream RAG methods retrieve the relevant knowledge from the textual chunks, which ignore the structured information and inter-relations of these textual chunks. 
To mitigate this limitation, Graph RAG~\citep{hu2024grag} and KG-RAG~\citep{sanmartin2024kg, xu2024retrieval, linders2025knowledge} are proposed.
Instead of retrieving the knowledge from textual chunks, Graph RAG directly retrieves the relevant knowledge from graph data. 
GRAG~\citep{hu2024grag} retrieves the top-$k$ relevant subgraphs from the textual graph \eat{that is indexed based on PLMs}and then integrates the retrieved subgraphs with the query by aggregating and aligning the graph embeddings with text embeddings based on GNNs.  
LEGO-GraphRAG~\citep{cao2024lego} decomposes the retrieval process into subgraph-extraction, path-filtering, and path-refinement modules, thereby improving the reasoning capabilities of LLMs with retrieved knowledge. 
KG$^2$RAG~\citep{zhu2025knowledge} retrieves the relevant subgraph from KG \eat{by leveraging BFS search over KGs}and expanding the textual chunks with the retrieved KG\eat{, and then incorporates the expanded chunks with the prompt} for augmenting the generation.

\subsubsection{Aligning with Complex QA}
In the approaches of synthesizing LLMs and KGs for QA, where KG plays the role of background knowledge, the retrieved relevant knowledge from the factual knowledge graph can reconcile knowledge conflicts from the multiple documents for multiple-doc QA~\citep{wang2024knowledge}. 
The cross-modal reasoning \eat{over the retrieved knowledge} can facilitate the cross-modal interaction and alignment for multi-modal QA~\citep{suri2024visdom}. 
Additionally, the question decomposition of multi-hop QA can be augmented by fusing the knowledge from LLMs and KGs, which further facilitates iterative reasoning for generating the accurate final question~\citep{saleh2024sg, cao2024lego}. 
The RAG~\citep{roy2024learning} and KG-RAG~\citep{sanmartin2024kg} can also improve the capabilities of LLMs in understanding the user's interactions for generating accurate answers for conversational QA. 
However, the key technical challenge behind this methodology is \textit{how to retrieve the relevant knowledge from large-scale KGs and then effectively fuse with LLMs without inducing knowledge conflicts?}

\subsection{KGs as Reasoning Guidelines} \label{sec:kgasrg}
KGs can provide reasoning guidelines for LLMs to access precise knowledge from \eat{and logical connections between} factual evidence based on reasoning.
Recent methods (\textbf{comparison in Table~\ref{tab:summary-comparsion-kg-as-rg}}) for integrating the reasoning of KG and LLMs can be classified into three categories. 
\subsubsection{Offline KG Guidelines} \label{subsec:kgasrg1}
In offline KG guidelines, KG supplies potential paths or subgraphs before the reasoning process by the LLM, from which the LLM selects the most relevant path for reasoning.
For instance, SR~\citep{zhang2022subgraph} trains a subgraph retriever that operates independently from the downstream reasoning process, while Keqing~\citep{wang2023keqing} decomposes complex questions using predefined templates, retrieves candidate entities and triples from a KG.
EtD~\citep{liu2024explore} initially uses a GNN to identify promising candidates and extracts relevant fine-grained knowledge pertinent to the questions, and then creates a knowledge-enhanced multiple-choice prompt to guide LLM for generating the final answer.
Recent studies have explored the application of novel formats of guidelines.
GCR~\citep{luo2024graph} first converts a KG into a KG-Trie and then develops a graph-constrained decoding and a lightweight LLM to generate multiple reasoning paths and candidate answers.
KELDaR~\citep{li2024framework} introduces the question decomposition and atomic retrieval modules to extract implicit information\eat{of reasoning as guidelines for atomic retrieval on KG} and retrieves the relevant subgraphs from KG to augment LLM for QA.

\subsubsection{Online KG Guidelines} \label{subsec:kgasrg2}
This paradigm emphasizes that the guidance of the KG directly involves the LLM's reasoning process, rather than merely relying on existing static knowledge. 
For example, Oreo~\citep{hu2022empowering} uses a contextualized random walk across a KG and conducts a single step of reasoning through the specific layers.
KBIGER~\citep{du2022knowledge} considers the ($k$-1)-th reasoning graph to construct the $k$-th instruction for reasoning and corrects erroneous predictions of intermediate entities.
LLM-ARK~\citep{huang2023llm} approaches reasoning tasks as sequential decision-making processes and employs Proximal Policy Optimization (PPO) for optimization.
ToG~\citep{sun2023think} allows LLMs to iteratively perform beam search over KGs, by which the most promising reasoning paths and the most likely reasoning outcomes are generated.
In contrast, ToG-2~\citep{ma2025think} utilizes entities as intermediaries to guide LLMs toward precise answers based on iterative retrieval between documents and KGs.
KG-CoT~\citep{zhao2024kg} leverages external KGs to generate reasoning paths for joint reasoning of LLMs and KGs to enhance the reasoning capabilities of LLMs for QA.
To identify the relevant KG subgraphs, HippoRAG~\citep{jimenez2024hipporag} integrates multip-hop reasoning with single-step multi-hop knowledge retrieval.

\subsubsection{Agent-based KG Guidelines} \label{subsec:kgasrg3}
\begin{figure}[ht]
    \centering
    \includegraphics[width=1.0 \linewidth]{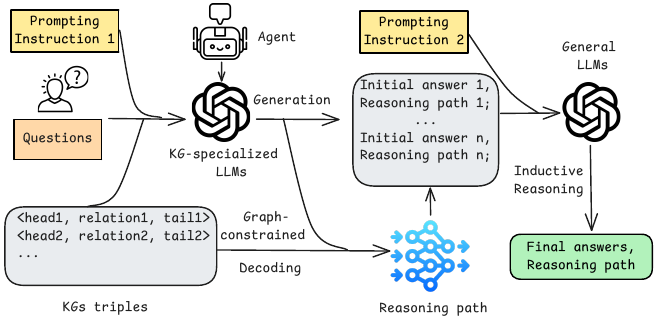}
    \caption{KGs as Reasoning Guidelines.}
    \label{fig:4}
\end{figure}

KGs can also be integrated into the reasoning process of LLMs as a component within an Agent system, as shown in Figure~\ref{fig:4}. 
This integration allows the Agent to leverage structured knowledge for augmenting the decision-making and problem-solving capabilities of LLMs.
KG-Agent~\citep{jiang2024kg} is a multifunctional toolbox integrating LLMs with a KG-based executor and a knowledge memory system that autonomously selects tools and updates the memory to enhance the reasoning of LLMs over KGs.
ODA~\citep{sun2024oda} incorporates KG reasoning capabilities through a global observation approach, which improves reasoning abilities by employing a cyclical paradigm of observation, action, and reflection.
GREASELM~\citep{zhang2022greaselm} effectively integrates encoded representations from LMs and GNNs by introducing several modality interaction layers to seamlessly blend structured knowledge with language contexts.
PoG~\citep{chen2024plan} integrates reflection and self-correction mechanisms to adaptively explore the reasoning paths over KG via an LLM agent, and then augments the LLM in complex reasoning and question answering based on the retrieved knowledge paths.
ATOMR~\citep{xin2025atomr} leverages LLMs as a reasoning agent to retrieve and incorporate the knowledge across multiple knowledge sources for augmenting the reasoning capability of LLM in knowledge-intensive question answering. 

\subsubsection{Aligning with Complex QA}
The approaches that incorporate KGs with LLMs can enable multi-hop and iterative reasoning over the factual KGs, further augmenting the reasoning capability of LLMs for complex QA.
The challenges of knowledge retrieval and conflicts across modalities and knowledge sources, as well as complex knowledge reasoning and question answering in multi-document QA, multi-modal QA, and multi-hop QA, can be mitigated through joint reasoning over factual KGs and LLMs~\citep{suri2024visdom,liang2025reasvqa,qiao2024graphllm,xin2025atomr}. 
Moreover, the joint reasoning over the factual KGs and LLMs provides the logical inference chains and anchor for LLMs to generate explainable answers with clear evidence from factual KGs~\citep{zhao2024kg\eat{, christmann2023explainable}}.
Although the joint reasoning over the factual KGs and LLMs can address the challenges of complex QA well, joint KG–LLM reasoning remains inefficient because large-scale graph traversal is computationally intensive and time-consuming.
Moreover, the reasoning capabilities of KGs mainly depend on the completeness and knowledge coverage of KGs, where the incomplete, inconsistent, and outdated knowledge from KGs might induce noise or conflicts.
The main challenge lies in \textit{how to improve the reasoning efficiency over the large-scale graph and reasoning capabilities under incomplete KG?}

\subsection{KGs as Refiners and Validators} \label{sec:kgasrv}
\begin{figure}[ht]
    \centering
    \includegraphics[width=1.0 \linewidth]{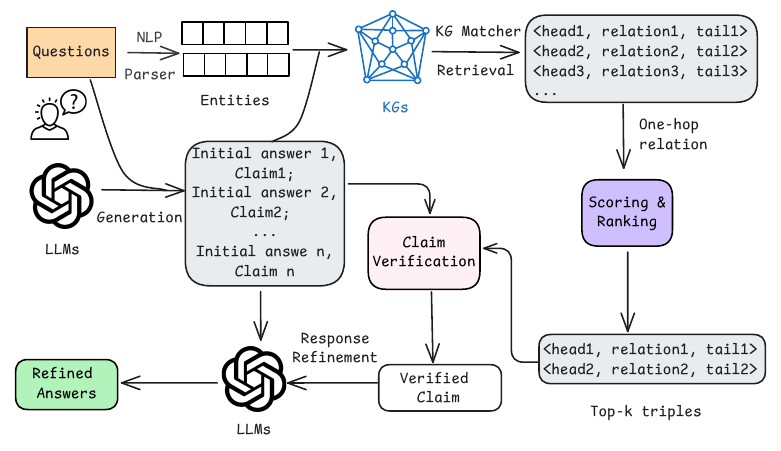}
    \caption{KGs as Refiners and Validators.}
    \label{fig:5}
\end{figure}

The factual evidence from KGs enables LLMs to refine and verify \eat{the correctness and relevancy of}the intermediate answers, as shown in Figure~\ref{fig:5}. 
In these methods (\textbf{comparison in Table~\ref{tab:summary-comparsion-kg-as-rav}}), KGs act as refiner and validator.
\subsubsection{KG-Driven Filtering and Validation} \label{subsec:kgasrv1}
KGs enhance the accuracy and reliability of LLM outputs by filtering and validating candidate answers with structured and verified knowledge. 
For instance, ACT-Selection~\citep{salnikov2023answer} filters and re-ranks answer candidates based on their types extracted from Wikidata.
Q-KGR~\citep{zhang2024question} improves the reasoning capabilities of LLMs by filtering out the irrelevant knowledge based on the ranking of the relevance score between the question and knowledge. 
KGs can improve the factual accuracy of generated answers, as demonstrated by KG-Rank~\citep{yang2024kg}, which integrates medical KGs with re-ranking techniques to enhance the credibility of generated responses. 
Moreover, KGR~\citep{guan2024mitigating} autonomously extracts and validates factual statements, significantly boosting performance on factual QA.

\subsubsection{KG-Augmented Output Refinement} \label{subsec:kgasrv2}
Integrating KGs with LLMs is essential to refining the outputs of LLMs for greater clarity and accuracy. EFSUM~\citep{ko2024evidence} employs LLM as a fact summarizer to generate relevant summaries from KGs, thereby improving performance in zero-shot QA. InteractiveKBQA~\citep{xiong2024interactive} enables iterative knowledge interactions, allowing LLMs to generate logical forms and refine outputs based on user feedback. LPKG~\citep{wang2024learning} fine-tunes LLMs with KG-derived planning data to enhance the planning capabilities of LLMs in sophisticated reasoning for complex QA.

\subsubsection{Aligning with Complex QA}
The approaches that leverage the retrieved factual evidence from KGs for refinement and validation are designed to augment the capability of LLMs in understanding user interactions and verifying the intermediate reasoning for multi-hop QA~\citep{chen2024llm} and conversational QA~\citep{xiong2024interactive},  respectively. 
However, the common case is that the knowledge from KGs is not available or not enough to verify the intermediate results, since the factual knowledge from KGs might be incomplete~\citep{zhou2025evaluating}.
In particular, the knowledge conflicts between the intermediate answer and KG facts might induce irrelevant results due to the poorly verified intermediate results. 
Meanwhile, the refinement and validation of results largely depend on the correctness, timeliness, and completeness of factual knowledge in KGs.
Thereby, the main challenges of this approach lie in \textit{how to handle the knowledge conflict between intermediate answers and KG facts and incrementally update KGs to ensure factual knowledge in KGs is up-to-date and correct?}

\subsection{Advancements} \label{sec:advance}
The advanced methods can be categorized into two subcategories: hybrid methods and optimization.
\subsubsection{Hybrid Method} \label{subsec:hybridm}
In addition to acting as a single role, the approaches (\textbf{comparison in Table~\ref{tab:summary-comparsion-hybrid-approah}}) where KGs serve multiple roles when synthesizing with LLMs for QA have been investigated. 
For instance, KG-RAG~\citep{sanmartin2024kg} introduces the Chain-of-Explorations (CoE) to rank and obtain the top-$k$ nodes or relationships based on vector similarity, by which the most relevant knowledge is selected and then fed into LLMs for the final answer. 
LongRAG~\citep{zhao2024longrag} retrieves the top-$k$ most relevant chunks based on a hybrid retriever and analyzes their relevance with the query by introducing a CoT-guided filter. 
Furthermore, the KG-augmented prompting is introduced to LLMs for augmenting the generation of the final answer. 
In KG-Rank~\citep{yang2024kg}, multiple ranking methods are introduced to refine the retrieved triples for augmenting the reasoning with the most relevant knowledge.
FRAG~\citep{zhao2024frag} introduces reasoning-aware and flexible-retrieval modules to retrieve reasoning paths from KGs, thereby guiding and augmenting LLM for efficient reasoning and answer generation.
KGQA~\citep{ji2024retrieval} combines the CoT-based prompting with graph retrieval to improve the retrieval quality and the multi-hop reasoning capability of LLMs in QA.

\subsubsection{Optimization} \label{subsec:optimization}

To mitigate the low efficiency and high computing costs of existing methods in synthesizing LLMs and KGs for complex QA, several optimization techniques (\textbf{extension in Appendix \S~\ref{sec:appendix-extended-optimization}}) are proposed to improve the efficiency of synthesizing LLMs and KGs. 
{\bf(1)} \textit{Index-based optimization.} It aims to accelerate the process of learning embeddings and vector retrieval for given questions and knowledge context. 
For instance, PG-RAG~\citep{liang2024empowering} proposes dynamic and adaptable knowledge retrieval indexes based on LLMs that can effectively handle complex queries and improve the overall performance of RAG systems in QA tasks.
{\bf(2)} \textit{Prompt-based optimization.} It mainly enhances the quality of the prompts and facilitates the knowledge fusion based on prompt engineering.
For example, KGP~\citep{wang2024knowledge} proposes a KG prompting approach to enhance the prompt for LLMs and optimize the knowledge retrieval by introducing the KG construction module and LLM-based graph traversal agent.
{\bf(3)} \textit{Cost-based optimization.} It aims to minimize computation costs by reducing the number of calls to LLMs and accelerating knowledge retrieval. 
In particular, SPOKE KG-RAG~\citep{soman2024biomedical} proposes a token-based optimized KG-RAG framework that integrates explicit and implicit knowledge from KG with LLM to enhance LLMs for cost-effective QA. 

\subsubsection{Aligning with Complex QA}
The hybrid method that involves the multiple roles of KG when synthesizing LLMs and KGs can address the limitations and challenges of complex QA in knowledge interactions and complex reasoning across modalities and knowledge sources.
The combination of knowledge fusion, RAG, CoT-based reasoning, and ranking-based refinement is capable of accelerating the complex question decomposition~\citep{wang2023keqing} for multi-hop QA, enhancing the context understanding for conversation QA~\citep{roy2024learning}, facilitating the interactions across modalities for multi-modal QA~\citep{dong2024modality}, and improving the explainability of the generated answers~\citep{christmann2023explainable}.
However, the remaining challenge is \textit{how to achieve efficient vector indexing and search over large-scale KGs and maintain a balance between the cost and performance?} 

\section{Evaluation and Application} \label{sec:evaluation-application}
We summarize the metrics \& dataset and showcase the related applications and demonstrations.
\subsection{Evaluation and Benchmark Dataset}
The metrics and benchmark datasets are summarized and compared (\textbf{details in Appendix \S \ref{subsec:appendix-evaluation}}).  

\spara{Metrics.} The metrics for evaluating the approach in synthesizing LLMs with KGs for QA are summarized: {\bf(1)} \textit{Answer Quality (AnsQ)};  {\bf(2)} \textit{Retrieval Quality (RetQ)}; {\bf(3)} \textit{Reasoning Quality (ReaQ)}.

\spara{Benchmark Dataset.} The recent benchmark datasets are summarized and compared (\textbf{details in Table~\ref{tab:benchmark-dataset-comparison}}) with the focus on \textit{Answer Quality}, \textit{Retrieval Quality}, and \textit{Reasoning Quality}. 

\subsection{Applications}
We showcase the industrial and scientific applications and demonstrations (\textbf{details in Appendix \S \ref{subsec:appendix-application}}) in synthesizing LLMs with KGs for QA.

\section{Open Challenges and Opportunities} \label{sec:open-challenges-opporitunities}
Even though the synthesis of LLM and KG can take advantage of their strengths and be mutually beneficial for complex question answering, the current LLM+KG QA systems may still meet challenges.
We summarize the challenges by highlighting the opportunities and discussing the future directions.

\spara{Scaling to both Effectiveness and Efficiency.} 
LLM+KG systems retrieve the factual knowledge and perform multi‐hop reasoning under tight latency and memory budgets. Three bottlenecks are emerging:
\textbf{(1) Structure‐aware retrieval}: Vanilla dense or sparse retrieval treats a KG as an unordered triple, thereby discarding topological cues that are vital for pruning the search space~\citep{tian2025systematic}. Hierarchical graph partitioning, dynamic neighbourhood expansion, and learned path‐prior proposal networks are promising ways to expose structure to the retriever while keeping the vector index sublinear.
\textbf{(2) Amortized reasoning}: Current retrieval and prompting pipelines repeatedly query the KG for every Beam search or CoT step. Caching subgraphs, reusing intermediate embeddings, and exploiting incremental‐computation friendly hardware can mitigate the quadratic blow‐up of iterative reasoning.
\textbf{(3) Lightweight answer validation}: Most guardrails rely on LLMs, while probabilistic logic programs and bloom filter sketches with KG-based fact-checking offer a lightweight solution. An opportunity is to design a retriever and validator that estimates the uncertainty of retrieval results and then guides the selective execution of validation.

\spara{Knowledge Alignment and Dynamic Integration.} 
Once a KG snapshot is injected into an LLM, it starts to become outdated, just like real‐world KGs usually involve adding new entities, deleting relations, and resolving contradictions.  Future work should focus on:
\textbf{(1) Quantify alignment}: We lack metrics that score not only semantic overlap but also \emph{structural} compatibility between parametric knowledge in the LLM and symbolic knowledge in the KG.  Contrastive probing with synthetic counterfactuals or topology‐aware alignment losses may fill this gap.
\textbf{(2) Facilitate real‐time updates}: Parameter‐efficient tuning (e.g.\ LoRA modules keyed by graph deltas) and retrieval‐time patching (streaming KGs with temporal indices) are early steps toward \emph{stream‐time} knowledge alignment.
\textbf{(3) Detect and resolve conflicts}: Bayesian trust networks, source‐aware knowledge distillation, and multi‐agent debate protocols can estimate and reconcile confidence scores across modalities and sources.  Incorporating these into the decoding objective is an open challenge with high pay‐off.

\spara{Explainable and Fairness-Aware QA.} 
The scale of LLMs poses challenges to explainability and fairness in QA. While integrating KGs offers a path toward interpretable reasoning, it also introduces computational challenges and fairness concerns. Future work may consider the following directions:
\textbf{(1) Reasoning over subgraphs}: Retrieving subgraphs from large-scale KGs is computationally expensive and often results in overly complex or incomprehensible explanations. Structure-aware retrieval and reranking methods should be employed to identify subgraphs consistent with the gold subgraphs. Furthermore, CoT-based prompting can be used to guide LLMs in generating explicit reasoning steps grounded in the retrieved subgraphs.
\textbf{(2) Fairness-aware knowledge retrieval}: 
LLMs can capture social biases from training data, but KGs may contain incomplete or biased knowledge. 
As a result, the fairness concerns remain in RAG~\citep{wu2024does}. 
Incorporating fairness-aware techniques into KG retrieval (e.g., reranking based on bias detection) and integrating them with counterfactual prompting can mitigate bias.
\textbf{(3) Explainable conversational QA}: Single-turn QA restricts the exploration of diverse perspectives and limits the reasoning processes. 
Developing conversational QA with retrieval strategies can dynamically detect and adjust knowledge biases and further improve the explainability of the RAG-based QA system through multi-turn user interactions.

\section{Conclusion}
This survey has systematically examined the synthesis of LLMs and KGs in question answering, presenting a novel taxonomy that categorizes methodologies based on QA types and the role of KGs. 
Our analysis highlights the strengths and limitations of current approaches, emphasizing the potential opportunities of leveraging the KGs to augment LLMs to overcome challenges such as hallucinations, limited reasoning capabilities, and knowledge conflicts in complex QA scenarios. 
Despite significant advancements, several remaining challenges include efficient knowledge retrieval, dynamic knowledge integration, effective reasoning over knowledge at scale, and explainable and fairness-aware QA. 
Future research should focus on developing adaptive frameworks that dynamically integrate up-to-date knowledge with LLMs, as well as establishing efficient methods for scaling reasoning, explainability, and fairness.

\section*{Limitations}
This survey covers the taxonomy and survey of the recent advancements in synthesizing LLMs and KGs for QA, as well as the discussion of its challenges and opportunities. However, we are aware that this survey may miss some newly released works due to the rapid expansion of works on this topic.
Moreover, the survey mainly highlights the alignments between the recent methodologies of incorporating LLMs and KGs for QA and the challenges of the various complex QA tasks, while these taxonomies from different perspectives are non-exclusive, and the overlap between the two taxonomies may arise. 
Furthermore, this survey underemphasizes the quantitative and experimental evaluation of different methodologies due to the various implementation details, the diversity of the benchmark datasets, and non-standardized evaluation metrics.

\section*{Acknowledgments}
This work was supported by the Novo Nordisk Foundation grant (NNF22OC0072415), the National Natural Science Foundation of China (U23B2057, 62176185, 62376058), and the GOBLIN COST Action (CA23147). 

\bibliography{custom}
\appendix

\section{Related Survey and Paper Selection}
\subsection{Related Survey} \label{subsec:appendix-related-survey}
Previous surveys have drawn the roadmap of unifying LLMs+KGs~\citep{pan2024unifying}, discussed the opportunities and challenges~\citep{pan2023large} of leveraging LLMs for knowledge extraction, ontologies and KGs constructions, summarized the integration and synthesis paradigms of LLMs and KGs~\citep{kau2024combining, ibrahim2024survey}, and gave an overview of the knowledge injection methods between LLMs and domain-specific knowledge~\citep{song2025injecting}.
Additionally, the existing surveys give an overview of multilingual knowledge graph question answering ~\citep{perevalov2024multilingual}, review temporal knowledge graph QA \citep{su2024temporal}, complex QA~\citep{daull2023complex}, and discuss search engines, KGs and LLMs from the perspective of user information seeking for QA \citep{hogan2025large}.
The comparative analysis of these surveys across LLMs, KGs, LLMs+KGs, GraphRAG, and QA is given in Table~\ref{tab:survey-comparsion}.
\FloatBarrier
\begin{table*}
    \footnotesize
    \centering
    \begin{tabular}{lccccc}
        \toprule
        \textbf{Survey} & \textbf{KGs} & \textbf{LLMs} & \textbf{LLMs+KGs} & \textbf{Graph RAG} & \textbf{QA} \\
        \midrule
        Unifying LLMs and KGs: Roadmap~\citep{pan2024unifying}  & \CheckmarkBold &  \CheckmarkBold & \CheckmarkBold & \CheckmarkBold  & \XSolid    \\  
        LLMs+KGs: Opportunities and Challenges~\citep{pan2023large}  & \CheckmarkBold &  \CheckmarkBold & \CheckmarkBold & \CheckmarkBold & \XSolid    \\ 
        Combing KGs and LLMs~\citep{kau2024combining}  & \CheckmarkBold &  \CheckmarkBold & \CheckmarkBold  & \CheckmarkBold & \XSolid    \\ 
        mQAKG~\citep{perevalov2024multilingual}  & \CheckmarkBold &  \XSolid & \XSolid & \XSolid  & \CheckmarkBold     \\ 
        TKGQA~\citep{su2024temporal}  & \CheckmarkBold &  \XSolid & \XSolid & \XSolid  & \CheckmarkBold     \\ 
        LLMs, KGs, and SEs~\citep{hogan2025large}  & \CheckmarkBold &  \CheckmarkBold & \CheckmarkBold & \XSolid  & \CheckmarkBold     \\ 
        Complex QA~\citep{daull2023complex}  & \CheckmarkBold &  \CheckmarkBold & \XSolid & \XSolid  & \CheckmarkBold     \\ 
        Knowledge Injection~\citep{song2025injecting}  & \CheckmarkBold &  \CheckmarkBold & \XSolid & \XSolid  & \CheckmarkBold     \\ 
         \midrule
        {\rm[This work]} & \CheckmarkBold &  \CheckmarkBold & \CheckmarkBold & \CheckmarkBold  & \CheckmarkBold  \\
        \bottomrule
    \end{tabular}   
    \caption{Comparsion of Existing Survey Across LLMs, KGs, LLMs+KGs, GraphRAG, and QA. mKGQA: multi-lingual Question Answering for Knowledge Graphs, TKGQA: Temporal Knowledge Graph QA. \CheckmarkBold~Covered or Dicussed, \XSolid~Not Covered or Dicussed.}
    \label{tab:survey-comparsion}
\end{table*}

We summarize that the previous surveys on synthesizing LLMs and KGs for QA show limitations in terms of the scope and tasks. {\bf(1)} The scope of the surveys\eat{~\citep{pan2024unifying, pan2023large, kau2024combining, ibrahim2024survey}} is dedicated for general knowledge-intensive tasks, such as knowledge extraction, entity linking, KG construction and completion, and text generation, etc; {\bf(2)} the QA task in the surveys\eat{~\citep{perevalov2024multilingual, su2024temporal}} is limited to the close-domain QA over the knowledge graph; {\bf(3)} the survey\eat{~\citep{hogan2025large}} on integrating LLMs, KGs, and search engines for answering user's questions is from the perspective of user-centric information needs.

\subsection{Paper Selection}
We first retrieve research papers since 2021 by using Google Scholar and PaSa\footnote{\url{https://pasa-agent.ai/}} using the phrases ``knowledge graph and language model for question answering" and ``KG and LLM for QA", and extend the search scope of the benchmark dataset paper to 2016, and then screen and select them based on their relevancy and publication venue quality. 

\section{Taxonomy} \label{sec:appendix-taxonomy}
\subsection{Complex QA}
We classify the complex QA into the following categories according to the technical challenges.
\begin{itemize}[noitemsep]
    \item \textbf{Multi-document QA.} Multi-document QA (Multi-doc QA) retrieves and synthesizes relevant information from various sources to provide a comprehensive answer.
    \item \textbf{Multi-modal QA.} It refers to the QA over data and knowledge involving multiple modalities such as text, audio, images, video, etc.
    \item \textbf{Muti-hop QA.} It usually involves decomposing the complex question and generating the final answers based on multi-step and iterative reasoning over a factual KG. 
    \item \textbf{Conversational QA.} It involves user engagement in multi-turn QA to understand the given context, determine the final answers, and satisfy their information needs.
    \item \textbf{Explainable QA.} XQA (Explainable QA) not only gives the answers to the question but also provides explanations for the given answers.
    \item \textbf{Temporal QA.} It refers to the questions with temporal intent over the temporal KGs that have entities, relations, and associated temporal conditions.
\end{itemize}

\subsection{Approaches of Synthesizing LLMs and KGs for QA}
We categorize the methodology of synthesizing LLMs and KGs for QA based on the role of KGs.
\begin{itemize}[noitemsep]
    \item \textbf{KG as Background Knowledge.} When KGs are used as background knowledge to enhance LLMs for QA, the questions are parsed to identify the relevant subgraphs from KGs, which are then incorporated with LLMs based on knowledge integration and fusion.
    \item \textbf{KGs as Reasoning Guidelines.} KGs can serve as reasoning guidelines for LLMs for QA tasks by providing structured real-world facts. This factual knowledge and their reliable reasoning paths can improve the explainability of the generated answers. 
    \item \textbf{KGs as Refiners and Validators.} KGs can be treated as refiners and validators for LLMs in QA tasks, where LLMs can verify initial answers with factual knowledge and filter out inaccurate responses by integrating KG to ensure that the final responses are precise.
    \item \textbf{Hybrid Methods.} The hybrid methods involve a KG's multiple roles, namely, background knowledge, reasoning guidelines, refiner and validator, when synthesizing LLMs and KGs for QA. 
\end{itemize}

We summarize the strengths and limitations of each method of synthesizing LLMs and KGs for QA in Table~\ref{tab:taxonomy-comparsion} according to our previous taxonomy. 
\FloatBarrier
\begin{table*}
   \centering
   \footnotesize
   \resizebox{\linewidth}{!}{
   \begin{tabular}{llll}
       \toprule
       \textbf{Approach} & \textbf{Strength} & \textbf{Limitation}  & \textbf{KG Requirement} \\
       \midrule
       KG as Background Knowledge  & Broad Coverage & Static Knowledge & High Domain Coverage         \\  
       KG as Reasoning Guidelines  & Multi-hop Capabilities & Computational Overhead & Rich Relational Paths    \\ 
       KG as Refiners and Validator & Hallucination Reduction & Validation Latency & High Accuracy \& Recency         \\ 
       Hybrid Approach & Limitation Mitigation & High Computing Cost  & Dynamic Adaptation   \\ 
       \bottomrule
   \end{tabular}}
    \caption{Comparison of Approaches with Different Roles of KGs.}
   \label{tab:taxonomy-comparsion}
\end{table*}

\section{Summary and Alignment} \label{sec:appendix-summary-and-alignment}
\subsection{Summary Tables of Approaches}
The detailed summarization and comparison of the different categories of approaches in terms of main techniques, language model (LM), knowledge graph (KG), dataset, QA types, and evaluation metrics are given in the following tables.

\spara{KG as Background Knowledge.} Table~\ref{tab:summary-comparsion-kg-as-bgk} gives a summary and comparison of approaches in the category of KG as background knowledge.
\begin{table*}[ht]
   \centering
   \footnotesize
    \resizebox{\linewidth}{!}{
   \begin{tabular}{p{4cm}p{3cm}p{3cm}p{2.5cm}p{2cm}p{2cm}p{2.5cm}}
       \toprule
       \textbf{Methods} & \textbf{Techniques} & \textbf{LM(s)} & \textbf{KG(s)} & \textbf{Dataset(s)}  &\textbf{QA Type} & \textbf{Metric(s)} \\
       \midrule
        \textbf{InfuserKI}~\citep{wang2024infuserki} & Knowledge-based Fine Tuning  & Llama-2-7B & UMLS, Movie KG (MetaQA) & PubMedQA, MetaQA-1HopQA  & KGQA  & NR, RR, F1  \\
        \textbf{KnowLA}~\citep{luo2024knowla} & Knowledgeable Adaptation  & Llama2-7B, Alpaca2 & WordNet, ConceptNet, Wikidata  & CSQA, SIQA, BBH, WQSP, TriviaQA & MCQA, CBQA, TruthfulQA  & Acc, CE Score, BLEU, ROUGE   \\ 
        \textbf{KG-Adapter}~\citep{tian2024kg} & Parameter Efficient Fine Tuning \& Joint Reasoning  & Llama-2-7B-base, Zephyr-7B-alpha &ConceptNet, Freebase  & OBQA, CSQA, WQSP, CWQ  & KGQA, MCQA, OBQA, CWQ  & Acc, Hits@1  \\ 
        \textbf{GAIL}~\citep{zhang2024gail} & GAIL Fine-tuning  & Llama-2-7B, BERTa  & Freebase & WQSP, CWQ, GrailQA  & KGQA  & EM, F1, Hits@1  \\ 
        \textbf{GRAG}~\citep{hu2024grag} & Textual Graph RAG & Llma-2-7B  & WebQSP, ExplaGraphs  & GraphQA, WQSP & KGQA & F1, Hits@1, Acc \\
        \textbf{LEGO-GraphRAG}~\citep{cao2024lego} & Modular Graph RAG & Qwen2-72B, Sentence Transformer & Freebase & WQSP, CWQ, GrailQA & KBQA, CWQ & R, F1, Hits@1 \\
        \textbf{KG$^2$RAG}~\citep{zhu2025knowledge} & Graph-guided Chunks Expansion & Llama-3-8B & Dataset Inherent KGs & HotpotQA & Multi-hop QA & F1, P, R \\
        \textbf{KG-RAG}~\citep{xu2024retrieval} & Vector-based Subgraph Retrieval & GPT-4 & Self-constructed KGs & Curated Dataset & KGQA & MRR, Recall@K, NDCG@K, BLEU, ROUGE, METEOR \\
       \bottomrule
   \end{tabular}}
   \caption{Summary and Comparison of Methods in the Category of KG as Background Knowledge. NR: Newly-learned Rate, RR: Remembering Rate, CE Score: Cross Entropy Score, EM: Exact Match, MRR: Mean Reciprocal Rank, BLEU: Bilingual Evaluation Understudy, ROUGE: Recall-Oriented Understudy, NDCG: Normalized Discounted Cumulative Gain, METEOR: Metric for Evaluation of Translation with Explicit Ordering.}
   \label{tab:summary-comparsion-kg-as-bgk}
\end{table*}

\spara{KG as Reasoning Guidelines.} Table~\ref{tab:summary-comparsion-kg-as-rg} gives a summary and comparison of approaches in the category of KG as reasoning guidelines.
\begin{table*}[ht]
   \centering
   \footnotesize
    \resizebox{\linewidth}{!}{
   \begin{tabular}{p{4cm}p{3cm}p{3cm}p{2.5cm}p{2cm}p{2cm}p{2.5cm}}
       \toprule
       \textbf{Methods} & \textbf{Techniques} & \textbf{LM(s)} & \textbf{KG(s)} & \textbf{Dataset(s)}  &\textbf{QA Type} & \textbf{Metric(s)} \\
       \midrule
        \textbf{SR}~\citep{zhang2022subgraph} & Trainable Subgraph Retriever, Fine-tuning & RoBERTa-base & Dataset Inherent KGs & WQSP, CWQ  & KBQA  & Hits@1, F1  \\
        \textbf{KELDaR}~\citep{li2024framework} & Question Decomposition Tree, Atomic KG Retrieval   & GPT-3.5-Turbo, GPT-4-Turbo & Dataset Inherent KGs  & WQSP, CWQ & KGQA, Multi-hop QA & EM   \\ 
        \textbf{Oreo}~\citep{hu2022empowering} & Knowledge Interaction and Injection, KG Random Walk  & RoBERTA-base, T5-base & Dataset Inherent KGs, Wikidata  & NQ, WQ, WQSP, TriviaQA, CWQ, HotpotQA  & CBQA, OBQA, Multi-hop QA  & Acc  \\ 
        \textbf{KBIGER}~\citep{du2022knowledge} & Interative Instruction Reasoning  & LSTM based Pre-trained Model  & Dataset Inherent KGs, Freebase & WQSP, CWQ, GrailQA  & Multi-hop KBQA  & Hits@1, F1  \\ 
        \textbf{ToG}~\citep{sun2023think} & Beam-search base Retrieval, LLM Agent & GPPT-3.5-Turbo, GPT-4, Llama-2-70B-Chat Sentence Transformer & Dataset Inherent KGs, Freebase, Wikidata & CWQ, WQSP, GrailQA, QALD10-en, WQ  & KBQA, Open-Domain QA  & Hits@1 \\
        \textbf{ToG-2}~\citep{ma2025think} & Hibrid RAG, Knowledge-guided Context Retrieval & GPT-3.5-Turbo, GPT-4o, Llama-3-8B, Qwen2-7B & Wikipedia, Wikidata & WQSP, QALD10-en, AdvHotpotQA, HotpotQA, ToG-FinQA & Multi-hop KBQA \& Document QA, Domain-specific QA & Acc, EM, R, F1 \\
        \textbf{KG-CoT}~\citep{zhao2024kg} & CoT-based KG and LLM Joint Reasoning & GPT-4, GPT-3.5-Turbo, Llama-7B, Llama-13B & Dataset Inherent KGs, Freebase & WQSP, CWQ, SQ, WQ & KBQA, Multi-hop QA & Acc, Hit@K \\
        \textbf{KG-Agent}~\citep{jiang2024kg} & KG-Agent based Instruction Tuning & Dvinci-003, GPT-4, Llama-2-7B & NQ-Wiki, TQ-Wiki, WQ-Freebas & WQSP, CWQ, GrailQA & KGQA, ODQA & Hits@1, F1  \\
        \textbf{ODA}~\citep{sun2024oda} & ODA-based Knowledge Graph Retrieval & GPT-4, GPT-3.5 & Wikidata & QALD10-en & KBQA & Hits@1, Acc \\
        \textbf{GREASELM}~\citep{zhang2022greaselm} & Mint-based KG and LM Cross-modal Fusion and Pretraing & RoBERTA-Large, AristoRoBERTA, SapBERT, PubmedBERT & ConcepNet, UMLS, DrugBank & CQA, OBQA, MedQAt & Multiple-choice QA & Acc \\
       \bottomrule
   \end{tabular}}
   \caption{Summary and Comparison of Methods in the Category of KG as Reasoning Guidelines. EM: Exact Match, Acc: Accuracy, R: Recall.}
   \label{tab:summary-comparsion-kg-as-rg}
\end{table*}

\spara{KG as Refiners and Validators.} Table~\ref{tab:summary-comparsion-kg-as-rav} gives a summary and comparison of approaches in the category KG as refiners and validators.
\begin{table*}[ht]
   \centering
   \footnotesize
   \resizebox{\linewidth}{!}{
   \begin{tabular}{p{4cm}p{3cm}p{3cm}p{2.5cm}p{2cm}p{2cm}p{2.5cm}}
       \toprule
       \textbf{Methods} & \textbf{Techniques} & \textbf{LM(s)} & \textbf{KG(s)} & \textbf{Dataset(s)}  &\textbf{QA Type} & \textbf{Metric(s)} \\
       \midrule
        \textbf{ACT-Selection}~\citep{salnikov2023answer} & ACT-based Answer Selection and Ranking  & T5-Large-SSM & Wikidata & SQ, RuBQ, Mintaka & KGQA, CBQA, Multi-lingual KGQA  & Hit@1  \\
        \textbf{Q-KGR}~\citep{zhang2024question} & Question-guided KG Re-scoring, FNN-based Knowledge Injection  & FLAN-T5-XL, RoBERTa-Large, Llama-2-7B  & ConceptNet  & OBQA, PIQA & KGQA & Acc   \\ 
        \textbf{KG-Rank}~\citep{yang2024kg} & Similarity and MMR based Ranking  & GPT-4, Llama-2-7B, Llama-2-13B & UMLS, DBpedia & LiveQA, ExpertQA-Bio, ExpertQA-Med, MedQA & Domain-specific QA & ROUGE-L, BERTScore, MoverScore, BLEURT  \\ 
        \textbf{KGR}~\citep{guan2024mitigating} & Refine-then-Retrieve, Knowledge Truthfulness Verfication & GPT-4, Llama-2-7B, Vanilla Llama-2-7B, Transformer & CKG, PrimeKG & MedQuAD & Domain-specific QA & Truthfulness Score  \\ 
       \textbf{EFSUM}~\citep{ko2024evidence} & KG Fact Summarization, KG Helpfulness and Faithfulness Filter  & GPT-3.5-Turbo, Flan-T5-XL, Llma-2-7B-Chat  &  Dataset Inherent KGs (Freebase, Wikidata) &  WQSP, Mintaka & KGQA, Multi-hop QA & Acc \\
       \textbf{InteractiveKBQA}~\citep{xiong2024interactive} & Multi-turn Interaction for Observation and Thinking  & GPT-4-Turbo, Mistral-7B, Llama-2-13B & Freebase, Wikidata, Movie KG & WQSP, CWQ, KQA Pro, MetaQA & KBQA, Domain-specific QA & F1, Hits@1, EM, Acc \\
        \textbf{LPKG}~\citep{wang2024learning} & Planning LLM Tuning, Inference, and Execution & GPT-3.5-Turbo, CodeQwen1.5-7B-Chat, Llama-3-8B-Instruct & Dataset Inherent KGs (Wikidata), Wikidata15K & HotpotQA, 2WikiMQA, Bamboogle, MuSiQue, CLQA-Wiki & KGQA, Multi-hop QA & EM, P, R \\
       \bottomrule
   \end{tabular}}
   \caption{Summary and Comparison of Methods in the Category of KG as Refiners and Validators. Acc: Accuracy, EM: Exact Match, BLEURT: Bilingual Evaluation Understudy with Representations from Transformers.}
   \label{tab:summary-comparsion-kg-as-rav}
\end{table*}

\spara{Hybrid Approach.} Table~\ref{tab:summary-comparsion-hybrid-approah} gives a summary and comparison of hybrid approaches.
\begin{table*}[ht]
   \centering
   \footnotesize
   \resizebox{\linewidth}{!}{
   \begin{tabular}{p{4cm}p{3cm}p{3cm}p{2.5cm}p{2cm}p{2cm}p{2.5cm}}
       \toprule
       \textbf{Methods} & \textbf{Techniques} & \textbf{LM(s)} & \textbf{KG(s)} & \textbf{Dataset(s)}  &\textbf{QA Type} & \textbf{Metric(s)} \\
       \midrule
         \textbf{LongRAG}~\citep{zhao2024longrag} & Domain-specific Fine-Tuning for RAG \& CoT-guided Filter  & ChatGLM3-6B, Qwen1.5-7B, Vicuna-v1.5-7B, Llama-3-8B, GPT-3.5-Turbo, GLM-4 & Wikidata & HotpotQA, 2WikiMQA, MusiQue & KBQA, Multi-hop QA & F1  \\
        \textbf{SimGRAG}~\citep{cai2024simgrag} & Instruction Fine-Tuning for RAG with Filter & Llama-3-8B-Instruct, Gemma-2-27B-it & Wikipedia, PubMed & PubMedQA, BioASQm MedQA, MedMCQA, LiveQA, MedicationQA & Domain-specific QA, Multi-choice QA & Acc, Rouge-L, MAUVE, EM, F1 \\
        \textbf{KGQA}~\citep{ji2024retrieval} & KG-related Instruction Tuning with CoT Reasoning & Llama-2-7B-Chat, BGE-1.5-en-base & Dataset Inherent KGs & WQAP, CWQ & KGQA & Hits@1, F1  \\ 
        \textbf{KG-IRAG}~\citep{yang2025kg} & Incremental Retrieval and Iterative Reasoning & Llama-3-8B-Instruct, GPT-3.5-Turbo, GPT-4o-mini, GPT-4o & Self-constructed KGs & TFNSW & Temporal QA & EM, F1, HR, HAL  \\ 
        \textbf{PIP-KAG}~\citep{huang2025pip} & Parameteric Pruning for KAG  & Llma-3-8B-Instruct  &  Dataset Inherent KGs & CoConflictQA & KGQA, Multi-hop QA & EM, ConR, MR   \\ 
        \textbf{RAG-KG-IL}~\citep{yu2025rag} & Agent-based Incremental Learning and Knowledge Dynamic Update & GPT-4o & Self-constructed KGs & Curated Dataset from NHS& Domain-specific QA & AM, HVC  \\
        \textbf{CoT-RAG}~\citep{li2025cot} & KG-driven CoT Generation and Knowledge-aware RAG with Pseudo-program KGs & ERNIE-Speed-128K, GPT-4o-mini & Self-curated Pseudo-Program KGs & HotpotQA, CSQA, SIQA & KGQA, Multi-hop QA & RA, Robustness \\
       \bottomrule
   \end{tabular}}
   \caption{Summary and Comparison of Methods in the Category of Hybrid Approach. EM: Exact Match, HR: Hit Rate, HAL: Hallucination, ConR: Context Recall, MR: Memorization Ratio, AM: Accuracy Matching, HVC: Human-verified Completeness, RA: Reasoning Accuracy.}
   \label{tab:summary-comparsion-hybrid-approah}
\end{table*}

\subsection{Alignment of Approaches to Complex QA}
The alignment of the existing approaches in synthesizing LLMs and KGs to diverse complex QA is presented in Table~\ref{tab:alignment-approaches}.
\begin{table*}[ht]
    \footnotesize
    \centering
    \resizebox{\linewidth}{!}{
    \begin{tabular}{lcccccc}
        \toprule
        \textbf{Approach} & \textbf{Multi-doc QA} & \textbf{Multi-modal QA} & \textbf{Multi-hop QA} & \textbf{Conversatiional QA} & \textbf{XQA} & \textbf{Temporal QA} \\
        \midrule
         KG as Background Knowledge  & \CheckmarkBold &  \CheckmarkBold & \CheckmarkBold & \CheckmarkBold  & $\triangle$   & \CheckmarkBold    \\  
         KG as Reasoning Guidelines  & \CheckmarkBold &  \CheckmarkBold & \CheckmarkBold  & \XSolid & \CheckmarkBold   & \CheckmarkBold   \\ 
         KG as Refiners and Validator   & $\triangle$ &  \XSolid & $\triangle$ & \CheckmarkBold & \XSolid  & $\triangle$   \\
         Hybrid Methods & \CheckmarkBold &  \CheckmarkBold & \CheckmarkBold & \CheckmarkBold  & \CheckmarkBold  & \CheckmarkBold    \\ 
        \bottomrule
    \end{tabular}}
    \caption{Research Progress on Alignment of Existing Approaches of Synthesizing LLMs and KGs with Complex QA. \CheckmarkBold~Fully Investigated. $\triangle$~Partially Investigated. \XSolid~Not Yet Investigated.} 
    \label{tab:alignment-approaches}
\end{table*}

\section{Extended Optimization} \label{sec:appendix-extended-optimization}
The extended optimization techniques are summarized and discussed as follows.

\spara{Index-based Optimization.} It is a time-consuming and complex task to create vector indexes from long-range facts and retrieve the relevant knowledge from large-scale graphs.  
To address this issue, GoR~\citep{zhang2024graph} leverages GNN and BERT score-based objectives to optimize node embeddings during graph indexing.  
KG-Retriever~\citep{chen2024kg} leverages a hierarchical index graph to enhance knowledge correlations and improve information retrieval for efficient knowledge indexing.
NodeRAG~\citep{xu2025noderag} integrates the heterogeneous graphs and fine-grained retrieval with RAG by optimizing the indexing of graph structures.
DRO~\citep{shi2025direct} proposes a direct retrieval-agumentation method to directly estimate the distribution of document permutations from the selected model for jointly learning the selection model and generative LM model.

\spara{Prompt-based Optimization.} To facilitate the deep fusion between the retrieved knowledge from KGs and the internal knowledge of LLMs, several prompt-based optimization approaches are proposed.  
For instance, StraGo~\citep{wu2024strago} enhances the quality and stability of the prompts based on the insights and strategic guidance learned from the historical prompts by using in-context learning.
Meanwhile, several ranking strategies have been exploited to retrieve the top relevant knowledge from diverse knowledge bases, and the most relevant contexts are further fed to LLMs together with the prompts. 
For example, KG-Rank~\citep{yang2024kg} leverages re-ranking techniques based on the score measuring relevance and redundancy to rank the top-$k$ triples from KGs and then combine them with the prompt to generate the answers for QA. 
Similarly, KS-LLM~\citep{zheng2024ks} introduces the evidence sentence selection module to retrieve the most relevant evidence sentences based on the ranking of the Euclidean distance between the triples and each evidence sentence.
Instead of direct ranking of context-based relevance, BriefContext~\citep{zhang2024mapreduce} introduces the preflight check to predict the relevance ranking between the user query and the retrieved documents, which is divided into multiple chunks for multiple RAG subtasks based on the map-reduce strategy. 
To leverage the multi-source knowledge for RAG-based QA, QUASAR~\citep{christmann2024rag} enhances RAG-based QA by effectively integrating unstructured text, structured tables, and knowledge graphs for evidence retrieval and re-ranking and filtering the relevant evidence from the retrieved evidence.

\spara{Cost-based Optimization.} To implement efficient and effective knowledge probing of LLMs, GLens~\citep{zheng2024kglens} initially leverages the Thompson sampling strategy to measure the alignment between KGs and LLMs for addressing the knowledge blind spots of LLMs, and then designs a graph-guided question generator to convert KGs to text, together with a sampling strategy on the parameterized KG structure to accelerate KG traversal. 
Similarly, Coke~\citep{dong2024cost} minimizes the calls of LLMs for KGQA by introducing a cluster-level Thompson sampling to formulate the accuracy expectation and an optimized context-aware policy to distinguish the expert model based on question semantics.
LMQL~\citep{beurer2023prompting} minimizes the call of LLMs by generating an efficient inference procedure based on the LMP (Language Model Programming) constraints and control flow. 
It differs from the above work in that CGPE~\citep{tao2024cluegoptimizing} optimizes the knowledge retrieval based on clue-guided path exploration and information matching from knowledge bases to enhance the capabilities of LLMs for unfamiliar questions and reduce the costs of LLMs. 

To summarize, even if several optimizations and ranking strategies have been recently investigated to reduce the costs of graph retrieval, graph reasoning, and the length of the context of LLMs. However, the relevant subgraphs extraction, graph reasoning, and vector-based retrieval remain a computationally costly task.  

\section{Evaluations and Applications}
\label{sec:appendix-evaluation-applications}
The details of evaluation metrics, benchmark datasets, industrial and scientific applications, and demonstrations of synthesizing LLMs and KGs for QA are summarized and compared as follows.
\subsection{Evaluations} \label{subsec:appendix-evaluation}
\spara{Metrics.} The evaluation metrics in synthesizing LLMs with KGs for QA are: {\bf(1)} \textit{the metrics measuring the answer quality}, BERTScore~\citep{peng2024graph}, answer relevance (AR), hallucination (HAL)~\citep{yang2025kg}, accuracy matching, human-verficed completeness~\citep{yu2025rag}; {\bf(2)}
\textit{the metrics measuring the retrieval quality of RAG}, context relevance \citep{es2024ragas}, faithfulness score (FS)~\citep{yang2024kg}, precision, context recall~\citep{yu2024evaluation, huang2025pip}, mean reciprocal rank (MRR)~ \citep{xu2024retrieval}, normalized discounted cumulative gain (NDCG)~\citep{xu2024retrieval}; {\bf(3)} \textit{the metrics measuring the reasoning quality for multi-hop QA}, Hop-Acc~\citep{gu2024pokemqa}, reasoning accuracy (RA)~\citep{li2025cot}.

\spara{Benchmark Dataset.} The recent benchmark datasets are summarized and compared in Table~\ref{tab:benchmark-dataset-comparison} with the focus on the following evaluations. 
{\bf(1)} \textit{Answer Quality (AnsQ)}: the correctness of the generated answer with supported evidence and retrieved context in comparison to the ground-truth answer; {\bf(2)} \textit{Retrieval Quality (RetQ)}: the relevance of the provided or retrieved context in comparison to the human-validated relevant context; {\bf(3)} \textit{Reasoning Quality (ReaQ)}: the correctness of generated reasoning chains and intermediate steps that explain how the final answer is derived. 

\begin{table*}[ht]
   \centering
   \footnotesize
   \resizebox{\linewidth}{!}{
   \begin{tabular}{p{5cm}llllp{5.2cm}}
       \toprule
       \textbf{Dataset} & \textbf{Category}  & \textbf{AnsQ} & \textbf{RetQ}  & \textbf{ReaQ} & \textbf{Brief Description} \\
       \midrule
       \href{https://www.microsoft.com/en-us/download/details.aspx?id=52763}{WebQSP}~\citep{yih2016value}  & KBQA & \CheckmarkBold  & \CheckmarkBold  & $\triangle$  & Includes SPARQL queries for knowledge-based QA.     \\ 
        \href{https://zenodo.org/records/7655130}{BioASQ-QA}~\citep{krithara2023bioasq} & KBQA  & \CheckmarkBold & \CheckmarkBold & \XSolid & Includes exact and ideal answers of question over multi-doc summarization. \\ 
       \href{https://github.com/aannonymouuss/CAQA-Benchmark}{CAQA}~\citep{hu2024benchmarking} & KBQA &  $\triangle$ & \CheckmarkBold & $\triangle$ & Evaluates complex reasoning and attribution. \\
       \href{https://github.com/D3Mlab/cr-lt-kgqa}{CR-LT KGQA}~\citep{guo2024cr} & KGQA  & \CheckmarkBold & $\triangle$ & \CheckmarkBold & Supports long-tail entities and commonsense reasoning. \\
        \href{https://exaqt.mpi-inf.mpg.de/}{EXAQT}~\citep{jia2021complex} & KGQA & \CheckmarkBold & \CheckmarkBold & $\triangle$ & Supports temporal QA with multiple entities, predicates, and conditions.\\
        \href{https://github.com/jonathanherzig/commonsenseqa}{CommonsenseQA}~\citep{talmor2019commonsenseqa} & Multi-choice QA  & \CheckmarkBold & $\triangle$ & \XSolid & Includes questions with the correct answer and four distractor answers. \\
        \href{https://github.com/jind11/MedQA}{MedQA}~\citep{jin2021disease} & Multi-choice QA  & \CheckmarkBold & $\triangle$ & $\triangle$ & Medical multi-choice QA dataset with multilingual medical examination text. \\   
        \href{https://github.com/probe2/multi-hop/}{MINTQA}~\citep{he2024mintqa} & Multi-hop QA &\CheckmarkBold & \CheckmarkBold & \CheckmarkBold &  Supports LLMs evaluation on new and tail knowledge. \\
        \href{https://github.com/zhudotexe/fanoutqa}{FanOutQA}~\citep{zhu2024fanoutqa} & Multi-Hop QA & $\triangle$ & $\triangle$ & \CheckmarkBold & Includes multi-hop question over multi-document.\\
        \href{https://github.com/yale-nlp/M3SciQA}{M3SciQA}~\citep{li2024m3sciqa} & Multi-modal QA & \CheckmarkBold & \CheckmarkBold & \CheckmarkBold & QA over the context across multiple document. \\
        \href{https://github.com/linyongnan/OMG-QA}{OMG-QA}~\citep{nan2024omg} & Multi-modal QA &\CheckmarkBold &\CheckmarkBold & $\triangle$ & Evaluates the retrieval and reasoning of QA across modalities. \\
        \href{https://scienceqa.github.io/}{ScienceQA}~\citep{lu2022learn} & Multi-modal QA & \CheckmarkBold & \XSolid& \CheckmarkBold & Supports multi-choice questions across disciplines in science topics.\\
        \href{https://github.com/temptabqa/temptabqa}{TempTabQA}~\citep{gupta2023temptabqa} & Temporal QA & \CheckmarkBold & $\triangle$ &\CheckmarkBold & Supports temporal reasoning over semi-structured tables.\\
        \href{https://github.com/jannatmeem95/PAT-Questions}{PATQA}~\citep{meem2024pat} & Temporal QA & \CheckmarkBold & $\triangle$ & \CheckmarkBold & Provides Wikidata questions for Present-anchored temporal QA. \\
        \href{https://github.com/weiyifan1023/MenatQA}{MenatQA}~\citep{wei2023menatqa} & Temporal QA & $\triangle$ & $\triangle$ & \CheckmarkBold & Evaluates the temporal reasoning capability of LLMs. \\
        \href{https://github.com/AlibabaResearch/DAMO-ConvAI/tree/main/Loong}{Loong}~\citep{wang2024leave} & Long-context QA & \CheckmarkBold & \XSolid & $\triangle$ & QA over multi-doc with a relevant document for the final answer. \\
        \href{https://github.com/mikejqzhang/SituatedQA}{SituatedQA}~\citep{zhang2021situatedqa} & Open-retrieval QA & \CheckmarkBold & \CheckmarkBold & \XSolid & Includes correct question with temporal or geographical context.\\
        \href{https://github.com/datadotworld/cwd-benchmark-data}{ChatData}~\citep{sequeda2024benchmark} & LLM-KG QA  & \CheckmarkBold & \CheckmarkBold &\XSolid & LLMs+KGs QA over the enterprise SQL database. \\
        \href{https://github.com/AKSW/LLM-KG-Bench}{LLM-KG-Bench}~\citep{meyer2023developing} & LLM-KG QA & \CheckmarkBold & \XSolid & \XSolid & Evaluates capabilities of LLMs in knowledge graph engineering. \\
        \href{https://github.com/chen-zichen/XplainLLM_dataset}{XplainLLM}~\citep{chen2024xplain} & LLM-KG QA & \CheckmarkBold & $\triangle$ & \CheckmarkBold & Focuses on QA explainability and reasoning. \\
        \href{https://github.com/Y-Sui/OKGQA}{OKGQA}~\citep{sui2024can} & LLM-KG QA & \CheckmarkBold & \CheckmarkBold & \CheckmarkBold & Evaluates LLMs+KGs for open-ended QA. \\
        \href{https://github.com/HKUDS/MiniRAG/tree/main/dataset/LiHua-World}{LiHua-World}~\citep{fan2025minirag} & LLM-KG QA &\CheckmarkBold &\CheckmarkBold &\CheckmarkBold & Evaluates the capability of LLMs on multi-hop QA in the scenario of RAG.  \\
        \href{https://stark.stanford.edu/}{STaRK}~\citep{wu2024stark} & LLM-KG QA & \CheckmarkBold &\CheckmarkBold & \CheckmarkBold & Evaluates the performance of LLMs-driven RAG for QA.\\
        \href{https://github.com/OpenBMB/PIP-KAG}{CoConflictQA}~\citep{huang2025pip} & LLM-KG QA &\CheckmarkBold &$\triangle$ &\CheckmarkBold & Evaluates contextual faithfulness for QA in the scenario of KAG.  \\
        \href{https://huggingface.co/datasets/Askio/mmrag_benchmark}{mmRAG}~\citep{xu2025mmrag} & LLM-KG QA &\CheckmarkBold & \CheckmarkBold &\CheckmarkBold & Evaluates multi-modal RAG including QA dataset across text, tables, and KGs.  \\
        \href{https://github.com/THU-KEG/AtomR}{BlendQA}~\citep{xin2025atomr} & LLM-KG QA &\CheckmarkBold & $\triangle$ &\CheckmarkBold & Evaluates cross-knowledge source reasoning capabilities of RAG for QA.  \\
       \bottomrule
   \end{tabular}}
    \caption{Comparison of LLMs+KGs for QA Benchmark Dataset. \CheckmarkBold~The dataset was primarily designed to support this evaluation. $\triangle$~The dataset can be adapted for this evaluation, but this evaluation is not their main focus. \XSolid~The dataset does not support this evaluation.}
   \label{tab:benchmark-dataset-comparison}
\end{table*}

\subsection{Applications} \label{subsec:appendix-application}
We showcase the applications and demonstrations in synthesizing LLMs with KGs for QA. 

\spara{KAG}~\citep{liang2024kag} (by Antgroup)\footnote{\url{https://github.com/OpenSPG/KAG}} is a domain-knowledge augmented generation framework that leverages KGs and vector retrieval to bidirectionally enhance LLMs for knowledge-intensive tasks, such as QA.

\spara{PIKE-RAG}~\citep{wang2025pike} (by Microsoft)\footnote{\url{https://github.com/microsoft/PIKE-RAG}} is a specialized knowledge and rationale augmented generation system with a focus on extracting, understanding, and applying domain-specific knowledge to guide LLMs toward accurate responses.

\spara{GraphRAG-QA} (by NebulaGraph)\footnote{\url{https://github.com/wey-gu/demo-kg-build}} is an industrial demo of GraphRAG integrating several query engines for augmenting QA, NLP2Cypher-based KG query engine~\citep{wu2022nebula}, vector RAG query engine, and Graph vector RAG query engine.

\spara{MedRAG}~\citep{zhao2025medrag} (by Nanyang Technological University et al.)\footnote{\url{https://github.com/SNOWTEAM2023/MedRAG}} is a KG-elicited reasoning enhanced RAG-based healthcare copilot that generates diagnoses and treatment recommendations based on the input patient manifestations.

\spara{Fact Finder}~\citep{steinigen2024fact} (by Fraunhofer IAIS and Bayer)\footnote{\url{https://github.com/chrschy/fact-finder}} augments LLMs with the query-based retrieval from medical KG to improve the completeness and correctness of answers.

\spara{AprèsCoT}~\citep{hirdel2025aprescot} (by University of Waterloo) is a visualization tool for explaining the answers generated by LLMs with CoT-based path exploration and inference over KGs.

\end{document}